# Overlapping oriented imbalanced ensemble learning method based on projective clustering and stagewise hybrid sampling

Fan Li[#], Bo Wang[#], Pin Wang[*], Yongming Li[*]

School of Microelectronics and Communication Engineering, Chongqing University, Chongqing, 400044, China

**Abstract**: The challenge of imbalanced learning lies not only in class imbalance problem, but also in the class overlapping problem which is complex. However, most of the existing algorithms mainly focus on the former. The limitation prevents the existing methods from breaking through. To address this limitation, this paper proposes an ensemble learning algorithm based on dual clustering and stage-wise hybrid sampling (DCSHS). The DCSHS has three parts. Firstly, we design a projection clustering combination framework (PCC) guided by Davies-Bouldin clustering effectiveness index (DBI), which is used to obtain high-quality clusters and combine them to obtain a set of cross-complete subsets (CCS) with balanced class and low overlapping. Secondly, according to the characteristics of subset classes, a stage-wise hybrid sampling algorithm is designed to realize the de-overlapping and balancing of subsets. Finally, a projective clustering transfer mapping mechanism (CTM) is constructed for all processed subsets by means of transfer learning, thereby reducing class overlapping and explore structure information of samples. The major advantage of our algorithm is that it can exploit the intersectionality of the CCS to realize the soft elimination of overlapping majority samples, and learn as much information of overlapping samples as possible, thereby enhancing the class overlapping while class balancing. In the experimental section, more than 30 public datasets and over ten representative algorithms are chosen for verification. The experimental results show that the DCSHS is significantly best in terms of various evaluation criteria.

# 1 Introduction

Thanks to the rapid development of electronic science and technology, a large number of datasets from various fields are available to help people analyze and promote industrial development. Inevitably, the number of samples in one class is often much larger than that of other classes, a phenomenon known as "data imbalance"[1]. Imbalanced data widely exist in areas such as medical diagnosis, fault detection, and credit loans, and their classification has become a hotspot and a challenging problem in the field of machine learning.

As we know, the challenge of imbalanced learning lies not only in class imbalance problem, but also in the class overlapping problem which is complex [2-3]. However, most of the existing algorithms just pay attention to the class balancing, seldomly considering the overlapping problem, so the performance is limited. For example, although the direct over-sampling of the majority samples can realize the class balancing, but meanwhile the class overlapping will possibly become worse. Here is another example. Grouping samples into subset with same or similar number of majority and minority samples can realize class balancing. But simple grouping samples will make the majority and minority samples overlap each other within subset. It will lead to unsatisfactory classification accuracy. Besides, some resampling methods mainly under-sample the majority samples in the class overlapping region to improve the visibility of the minority classes in the region [4-8]. However, due to the low accuracy of the existing overlap region identification algorithms, this type of approach tends to lead to the loss of effective information of majority sample or the residual of overlapping redundant majority sample. The limitations above prevent the existing methods from breaking through. Therefore, it is very necessary to design an ensemble learning method which can solve the class imbalance while keeping the class overlapping better.

In order to solve the problem above, we have the following solution. As to the grouping of samples, we design joint clustering and dimensional reduction, thereby finding the grouped subsets with lowest overlapping under some dimension. As to the sampling, we accurately recognize the overlapping area, and simultaneously under-sample the majority samples and oversample the minority samples without losing useful information. As to balancing samples, we design transfer mapping mechanism (CTM) to obtain new samples with smaller size and higher robustness. Based on the innovations above, we can consider solve the class imbalance and class overlapping problems as well, different from the existing methods.

The main contributions of this paper are as follows:
1) According to the limitation of the grouping of subsets in the existing imbalance ensemble learning methods, we proposed a projective clustering combination strategy (PCC). The strategy can realize the joint clustering and dimensional reduction, thereby minimize the class overlapping within grouped subsets. cross-complete subsets (CCS) with high quality are obtained.
2) According to the limitation of existing oversampling and under-sampling methods, a hybrid sampling method (SHS) is proposed for samples balancing. Local overlapping region dynamic search method (LORDS) can recognize overlapping areas more comprehensively and accurately than existing related methods without over-recognition. Different from the existing algorithms, the SHS can effectively ensure the information redundancy of the under-sampled majority samples and the safety and effectiveness of the oversampled minority samples, thereby solving the class imbalance and class overlapping problems of each subset of CCS.
3) According to the balanced subsets, we further designed an envelope clustering transfer mapping mechanism (CTM) for sample transformation. By multi-layer transformation, the proposed CTM can reconstruct the processed subsets to obtain samples with lower class overlapping and higher quality, which can assist the training and prediction of subsequent classifiers.

4) Combining the methods above, we propose an ensemble learning method based on the dual clustering and stage-wise hybrid sampling for imbalanced learning. Compared with the existing related algorithms, DCSHS can solve the class imbalance and class overlapping problems simultaneously during every steps. In addition, another advantage of this algorithm is that it can adopt a non-direct loss of majority samples to learn the sample information as fully as possible while avoiding the negative impact of overlapping majority samples on the classification.

The rest of the paper is organized as follows: the related works of imbalanced data classification are summarized briefly in Section 2; Section 3 introduces the working principle and design flow of the proposed algorithm in this paper; Section 4 organizes the experiments and analyzes experimental results. Finally, discussion and conclusions are presented in Section 5 and 6 respectively.

## 2 Related works

In response to the two major problems common to imbalanced data mentioned above, class imbalance and class overlapping, researchers have successively proposed a large number of related algorithms. Among them, studies that address class imbalance are more comprehensive, while relatively few studies also take class overlapping into account.

In general, the existing methods can be divided into two categories, namely data-level methods and algorithm-level methods. Data-level methods are further divided into resampling methods, which include oversampling, undersampling and hybrid sampling, and feature methods, which mainly refer to feature selection [9]. Resampling methods attempt to rebalance the class distribution of data by resampling samples to fit traditional classification algorithms. Feature methods attempt to reduce the class overlapping of the dataset by feature engineering to improve its separability. Algorithm-level methods can be further divided into cost-sensitive class methods, ensemble learning methods. The core idea of all these algorithms is to improve the classification mechanism of existing algorithms to accommodate imbalanced data.

Oversampling methods is mainly used in class imbalance studies, which solve the class imbalance problem by increasing the number of minority samples through certain strategies, with the disadvantage that it may introduce redundant minority class information and disrupt the judgment of the classifier. The simplest oversampling method is to randomly replicate (RR) the minority samples for balance, but since a balanced dataset constructed in this way contains many identical samples, it may easily lead to overfitting. To overcome this problem, Chawla et al [10] proposed the classical synthetic minority class oversampling technique (SMOTE), whose main idea is to randomly interpolate between minority samples and their nearest neighbors, which greatly enhances generalization compared to RR. In response to the various shortcomings of SMOTE, a series of SMOTE improvement algorithms have been proposed by subsequent researchers such as Borderline-SMOTE [11], incremental synthetic balancing algorithm (ISBA) [12]. In addition, some researchers have tried to generate new samples more efficiently with the aid of clustering. For example, Douzas et al [13] proposed assigning larger oversampling weights to minority class regions with smaller cluster imbalance and smaller density of cluster minority classes, with the core idea of synthesizing more minority class instances for bounded and sparse minority class regions. Tao X et al [14] clustered the samples into two classes based on K-means and interpolated between minority classes with unchanged class characteristics, i. e., minority class safety regions. GAN networks have also been used in recent years for the generation of minority samples in imbalanced samples due to their high clarity and realism of the generated samples. For example, in the literature [15-16], adversarial generative network GANs are introduced to capture the true data distribution and thus generate more realistic and valid minority samples.

Undersampling methods have applications in both class imbalance and class overlapping, which is the opposite of the idea of oversampling to increase the minority class. It balances the dataset by reducing the number of majority samples through certain strategies, with the disadvantage that some useful majority samples may be eliminated, resulting in the loss of sample information. The simplest relevant method is random undersampling, that is, randomly selecting a certain number of majority samples for removal [17], which results in a large amount of blindness. Among the methods for targeted removal of redundant majority samples, clustering assistance is also effective. For example, Rayhan et al [18] balanced the training set by randomly removing a certain percentage of samples from each cluster after clustering the majority. Lin W C et al [19] directly used a clustering algorithm to cluster the majority into clusters with the number of minority classes and used the cluster centers or their nearest neighbors to form a balanced dataset with the minority. Different from the literature [18], another study [20] used the instance selection method, including the genetic algorithm, IB3 and drop3, to select more representative samples from each majority cluster. The core idea of all of the above methods is to take advantage of clustering to better retain the representative sample information while reducing the number of the majority.

When it comes to class overlapping problems, researchers usually choose to undersample the overlapping majority samples or combine it with the oversampling of minority samples due to its own limitation, which aims to enhance the visibility of the minority samples in the overlap region and thus improve the classification performance on imbalanced datasets. Obviously, this kind of method allows the classifier to better learn the recognition rate of minority samples in the overlap region while inevitably losing the information of majority samples in the overlap region, which is the main improvement of this paper. The core of this type of method is how to more accurately identify the overlapping class region in the training set and avoid overrecognition of the region, thus preventing additional information loss of useful majority samples when undersampling the majority samples in the region. Therefore, the main difference between related methods is reflected in the identification strategy for overlapping samples. For example, Das B et al [4] considered the majority samples from clusters where a minority class proportion is greater than a threshold value as the overlapping majority samples and eliminated them. Vuttipittayamongkol et al [5] clustered the samples into two clusters based on FCM and considered the majority in the minority representative cluster as the potential overlapping sample, and those samples whose affiliation with the cluster is greater than the threshold α-cut are identified as overlapping majority samples. Alejo et al [6] eliminated those majority samples whose graph neighbors are not all majority classes by creating a Gabriel graph for each sample. One study [7] performed two iterations based on KNN to search the overlapping region. The first iteration searches the majority samples in the common neighborhood of the minority samples, and the second iteration continues to search the public neighborhood majority samples of the majority samples in the first search, and the results of the two searches are the overlapping majority samples. On the basis of the literature [7], the authors additionally proposed three neighborhood-based overlapping region identification algorithms [8]. In method 1, all majority samples whose K-neighborhood contains minority samples are regarded as overlapping samples. Method 2 adds the condition that the neighborhood of the minority sample must contain the majority samples on the basis of method 1, i. e., if a majority sample and a minority sample appear in each other's K-nearest neighbors, the majority sample will be identified as an overlapping sample. Method 3 performs a neighborhood search for the minority samples based on KNN, and if a majority sample appears in the common majority sample neighborhood of any two minority classes at the same time, it is considered an overlapping sample. The method in the literature [7] is also an improvement on this method 3. In the experimental part, a comparative experiment is performed on the methods in the literature [5] and [7] and the LORDS proposed in this paper.

Hybrid sampling methods are a combination of oversampling and undersampling methods. This kind of method aims to design a certain strategy to retain as many valid majority samples as possible and generate as few valid minority samples as possible through a compromise approach. Thus, the shortcomings of each of the two types of methods are mitigated to a certain extent, and the distribution of the samples is effectively equalized while remaining as close as possible to the original samples to ensure the authenticity of the data. Different hybrid sampling algorithms have their own preferences for the selection of the removed majority samples and the generation of minority samples. Min Z et al [21] used SMOTE to generate minority samples and then used the Tomek-Link technique to undersample the noisy majority samples. Al Majzoub H et al [22] chose to undersample the noisy majority samples and synthesize the minority samples near the boundary line based on the clustering result of the minority. One study [23] adopted a strategy of undersampling the densely distributed regions of the majority and oversampling the boundary regions of the minority. Another study [24] used the misclassification rate when samples are classified using RF to guide M-SMOTE to synthesize minority classes and used edited nearest neighbor method (ENN) to remove noisy samples from the majority.

Feature selection methods are mainly applied in class overlapping related studies, which usually tries to improve the separability of data by selecting features with lower class overlapping. For example, Rakkeitwinai et al [25] proposed selecting relevant features based on the overlap degree of dimensional data, which overcomes the limitation that the commonly used feature reduction algorithm PCA does not consider the data in each dimension of the feature space. Fu G H et al [26] proposed a feature selection algorithm based on a sparse regularization technique aiming at minimizing the overlap between the majority and the minority.

Cost-sensitive learning methods are mainly used to deal with the class imbalance problem, which corrects the bias of the classifier to majority samples by assigning a higher misclassification cost to minority samples according to certain criteria. The drawback of this kind of method is that the individual parameters of the cost matrix used to penalize the misclassification are difficult to determine effectively by a sound mathematical derivation. The literature [27] proposed a hybrid attribute metric to penalize misclassification based on the Gini index and information gain metric in decision trees to construct a cost-sensitive hybrid attribute metric multiple decision tree approach. Another study [28] adaptively determined misclassification cost weights in each iteration based on the previous classifiers obtained during the boosting process by adaptively considering the different contributions of minority instances to the SVM classifier. A study [29] proposes two cost-sensitive models based on support vector data description (SVDD) to minimize classification costs while maximize classification accuracy.

Existing ensemble learning methods are also mainly used to solve the class imbalance problem with less regard to class overlapping. Researchers often construct several balanced subsets of data to train weak classifiers separately by some strategy and then integrate them to better combine the performance of each weak classifier and greatly improve the generalizability of the model. This class of methods is usually used with resampling class methods, and the main difference between relevant methods is the different strategies used to construct the balanced subsets. For example, Ofek et al [30], after clustering the minority classes, sampled the same number of majority samples around the clusters as the minority samples to construct multiple balanced subsets. Each subset is trained separately and then fused together with the weights computed by the inverse distance between the test sample and the classifier. Sun et al [31] obtained multiple balanced subsets by differential sampling with different sampling rates for majority samples and minority samples after SMOTE expansion. Zhao J et al [32] combined two data sampling methods and two basic classifiers in the framework of a boosting algorithm and assigned corresponding weights to each sampling method and each basic classifier so that they can complement each other. Niu K et al [33] sampled each cluster several times according to the number proportion of clusters obtained from the majority cluster and combined it with the minority samples into multiple balanced subsets. Finally, the remaining classifiers filtered by the average AUC value of the validation set classification are used for voting fusion of the test samples.

In addition to the above methods, there are other algorithm-level methods that are inconvenient to categorize. For example, in the literature that focuses solely on the class imbalance problem, the literature [34] optimizes the parameters of the RBF classifier structure and RBF kernel using PSO based on the criterion of minimizing the Leak-misclassification rate and uses them for the classification of balanced datasets after SMOTE balance. The literature [35] introduces the relative density of the K-nearest neighbor-based probability density estimation strategy in FSVM, which can better capture the information of prior data distribution and correct the skewed data problem during classification. In the literature [36], a feature selection method based on a complex distance metric was designed for imbalanced datasets to solve the problem of intraclass imbalance and interclass imbalance. The literature [37] incorporates Gaussian similarity controlled by the Marxian distance into a dual support-based TWSVM to enable it to better handle imbalanced data.

In the rest of the literature where class overlapping is also considered, Prachuabsupakij et al [38] proposed a novel approach that clusters samples into two classes based on the K-means clustering algorithm, switches each other's positive class and negative class, and then trains the classifiers for integration. The idea of this approach is to try to reduce the degree of overlap between the minority and the majority in each subset. Some researchers choose to separate the overlapping region from the training set for separated processing. For example, Vorraboot et al [39] proposed a hybrid algorithm to divide the data into non-overlapping regions, boundary regions, and overlapping regions separately; Ren et al[40] used the AdaBoost algorithm to classify the overlapping region recognized by fuzzy KNN. In addition, direct modification of the classifier recognition mechanism is a popular practice. For instance, the literature [41] introduced an interclass consistency criterion incorporating the degree of sample overlap in KNN to guide the classification of imbalanced datasets, and the literature [42] proposed a fuzzy support vector machine algorithm with improved decision boundaries that can divide the data space into soft overlapping regions and hard overlapping regions for classification.

In summary, through the existing study above, it can be found that these algorithms have achieved certain results, but the following problems still exist. Firstly, most of the current relevant classification algorithms target the class imbalance problem, while the research on the class overlapping problem is ignored. Secondly, the existing recognition algorithms of class overlapping samples generally have the problem of over- or under-identification, which easily leads to the loss of effective majority samples' information or the residual of invalid redundant majority samples' information in the subsequent under-sampling process. Thirdly, the existing class overlapping-related imbalanced algorithms have not yet solved the problem of "how to fully learn class overlapping sample information". Last but most important, the existing imbalanced learning algorithms did not consider the sample transformation to obtain high quality of new samples for imbalanced learning. Fourthly, the existing imbalanced learning algorithms only consider the original samples. If the quality of the original samples is low (high class overlapping, high class imbalance), the algorithms cannot be satisfying. Therefore, it is necessary to transform the original samples to the ones with higher quality.

To address the problems above, this paper proposes the DCSHS algorithm, which can alleviate the above problems to a certain extent. The detail on the proposed DCSHS algorithm is described in 'method' section (section 3).

# 3 The proposed method

The proposed algorithm DCSHS is divided into three major parts-PCC, SHS and CTM. Among them, PCC is designed to obtain high quality of majority class clusters and minority class clusters, which are combined to obtain cross-complete set to avoid the overlapping distribution problem and facilitate subsequent processing. SHS is used to perform de-overlap and balance operations on each subset of CCS to improve the visibility of overlapping minority samples and the sample separability of each subset. CTM can construct high-quality samples with lower class overlapping and richer single-sample information, which can be used to assist the learning and prediction of corresponding weak classifier. The core of DCSHS is that PCC can work with SHS and CTM to achieve non-absolute rejection of overlapping majority samples in the training set, thus ensuring that the training of subsets is not affected by the class overlapping problem and fully learning the information of overlapping sample in the training set.

## 3.1 Projective clustering combination strategy (PCC)

Each class of samples in an imbalanced dataset often has different clustering and feature distributions within each subset, and the clustering method is effective in the existing literature. Therefore, to better avoid the adverse effects of loose mixing of distributions common in imbalanced datasets, this paper proposes the PCC strategy for constructing a cross complete set to process the samples of different clusters and different distributions independently. PCC uses the Davies-Bouldin clustering validity index as a loss function to guide the selection process of the projection matrix dimension, thus obtaining more valid clustering results. The clustering process is performed separately for the majority and the minority, and the final subset is obtained by a two-by-two combination of each majority class cluster and minority class cluster.

Assuming the training set $\mathbf{X} = \{(\mathbf{x}_i, y_i)\}_{i=1}^{N_s}$, in which an arbitrary sample $\mathbf{x}_i = (x_i^1, x_i^2, ..., x_i^{N_d})$. PCC first centralizes all samples according to Eq. (1).

$$\mathbf{x}_i = \mathbf{x}_i - \frac{1}{N_s} \sum_{i=1}^{N_s} \mathbf{x}_i \tag{1}$$

Then the covariance matrix $\mathbf{C}$ of $\mathbf{X}$ is constructed as follows:

$$\mathbf{C} = \begin{bmatrix} \mathrm{cov}(\mathbf{X}^1, \mathbf{X}^1) & \mathrm{cov}(\mathbf{X}^1, \mathbf{X}^2) & \cdots & \mathrm{cov}(\mathbf{X}^1, \mathbf{X}^{N_d}) \\ \mathrm{cov}(\mathbf{X}^2, \mathbf{X}^1) & \mathrm{cov}(\mathbf{X}^2, \mathbf{X}^2) & \cdots & \mathrm{cov}(\mathbf{X}^2, \mathbf{X}^{N_d}) \\ \vdots & \vdots & & \vdots \\ \mathrm{cov}(\mathbf{X}^{N_d}, \mathbf{X}^1) & \mathrm{cov}(\mathbf{X}^{N_d}, \mathbf{X}^2) & \cdots & \mathrm{cov}(\mathbf{X}^{N_d}, \mathbf{X}^{N_d}) \end{bmatrix} \tag{2}$$

where $\mathbf{X}^i, i=1,2,...,N_d$ is the $i$-th dimensional data of $\mathbf{X}$, $cov(\cdot)$ denotes the covariance calculation operation. Suppose the projection dimension is $d$, the projection matrix $\mathbf{T}_d$ can be obtained by singular value decomposition of $\mathbf{C}$ according to Eq. (3).

$$\mathbf{C}t = \lambda t \tag{3}$$

$$\mathbf{T}_d = [t_1, t_2, ..., t_d] \tag{4}$$

where $\lambda$ and $t$ represent the eigenvalues and eigenvectors of $\mathbf{C}$, respectively, $t_i, i=1,2,...,N_d$ denotes the eigencolumn vector corresponding to the *i*-th largest eigenvalue. Therefore, the sample $\dot{\mathbf{X}}$ after dimension reduction of $\mathbf{X}$ is expressed as Eq. (5).

$$\dot{\mathbf{X}} = \mathbf{X} \cdot \mathbf{T}_d = \mathbf{X} \cdot [t_1, t_2, ..., t_d] \tag{5}$$

After the number of majority class clusters $NC_{maj}$ and the number of minority class clusters $NC_{min}$ are given by prior statistical knowledge, PCC takes $d$ as 1 to $N_d$ respectively, and transforms the two classes of $\dot{\mathbf{X}}$ through the corresponding projection matrix $\mathbf{T}_d$ to obtain the corresponding majority dimension-reduction samples $\dot{\mathbf{X}}_{maj}$ and minority dimension-reduction samples $\dot{\mathbf{X}}_{min}$. By observing the feature distribution of a large number of imbalanced datasets, we can know that for most datasets, the number of clusters of the majority and the minority is less than or equal to 3. Therefore, the general value range of parameter $NC_{maj}$ and integer $NC_{min}$ is $[1,3]$.

Then, based on K-means, $\dot{\mathbf{X}}_{maj}$ and $\dot{\mathbf{X}}_{min}$ are clustered according to the given number of clusters $NC_{maj}$ and $NC_{min}$. The K-means clustering algorithm is a fast and effective unsupervised clustering method [43]. After each clustering, the Davies-Bouldin clustering validity index (DBI), one of the classical clustering effectiveness, of the majority and the minority are calculated by Eq. (9) and expressed as $DB_{maj}$ and $DB_{min}$ respectively. As Eq. (7) shows, the dimension corresponding to the minimum $DB_{maj} + DB_{min}$ value is selected as the final target dimension $d_t$ in this paper, and its corresponding clustering result will be used as the input of SHS step in our algorithm. The DBI can measure the quality of the clustering. As the formula (9-10) shows, the minimization of the $DB_{maj} + DB_{min}$ can make the clusters apart as far as possible. After combination of the majority samples and minority samples, the two classes of clusters are as simple and centralized as possible. In other words, the class overlapping will become better after grouping of subsets.

$$DB(d) = \frac{1}{NC} \sum_{m=1}^{NC} \max_{n \neq m} \left( \frac{\frac{1}{N_m} \sum_{i=1}^{N_m} \|x_{m,i}(d) - \mu_m(d)\|^2 + \frac{1}{N_n} \sum_{j=1}^{N_n} \|x_{n,j}(d) - \mu_n(d)\|^2}{\|\mu_m(d) - \mu_n(d)\|^2} \right) \tag{6}$$

$$d_t = \arg \min_d DB_{maj}(d) + DB_{min}(d) \tag{7}$$

where $NC$ denotes the number of clusters clustered by the majority or the minority. After clustering the corresponding data reduced to $d$ dimension, $x_{m,i}(d), i=1,2,...,N_m$ and $x_{n,j}(d), j=1,2,...,N_n$ denote an arbitrary sample in the *m*-th and *n*-th cluster, $N_m$ and $N_n$ are the number of these two clusters, and $\mu_m$ and $\mu_n$ are the central samples of them. The pseudocode of the projective clustering strategy (PC) is shown in Algorithm 1.

**Algorithm 1:** *Projective clustering - PC*

**Input**: Dataset $\mathbf{X}$, the number of clusters $NC_{maj}, NC_{min}$.

**Output**: The target dimension $d_t$, the clustering results of $d_t$.

1. Centralize $\mathbf{X}$ via (1);
2. Construct the covariance matrix $\mathbf{C}$ via (2);
3. Calculate the eigenvalues $\lambda$ and the eigenvectors $t$ via (3);
4. **For** $d = 1:N_d$
5.       Obtain the corresponding projection matrix $\mathbf{T}_d$ of dimension $d$ via (4);
6.       Transform features of $\mathbf{X}$ to obtain $\dot{\mathbf{X}}$ via (5);
7.       Divide the majority class samples $\dot{\mathbf{X}}_{maj}$ and minority class samples $\dot{\mathbf{X}}_{min}$ in $\dot{\mathbf{X}}$;
8.       Cluster the datasets $\dot{\mathbf{X}}_{maj}$ and $\dot{\mathbf{X}}_{min}$ respectively based on K-means via (6-8);
9.       Calculate the index values $DB_{maj} + DB_{min}$ of current clustering result via (9,10);
10. **end for**
11. Find the target dimension $d_t$ corresponding to the smallest $DB_{maj} + DB_{min}$ value;
12. **Return** $d_t$ and its corresponding clustering result.

For the clustering results of the majority and the minority based on projective clustering, we combine them to obtain all cross-complete subsets for subsequent processing. In order to better show the core idea of PCC strategy, we draw the schematic diagram of PCC strategy as shown in Fig. 1. Due to the characteristics of this strategy, we call the constructed set cross-complete set. As can be seen from Fig. 1, CCS has three satisfying characteristics: simple class distribution, intersectionality and completeness.

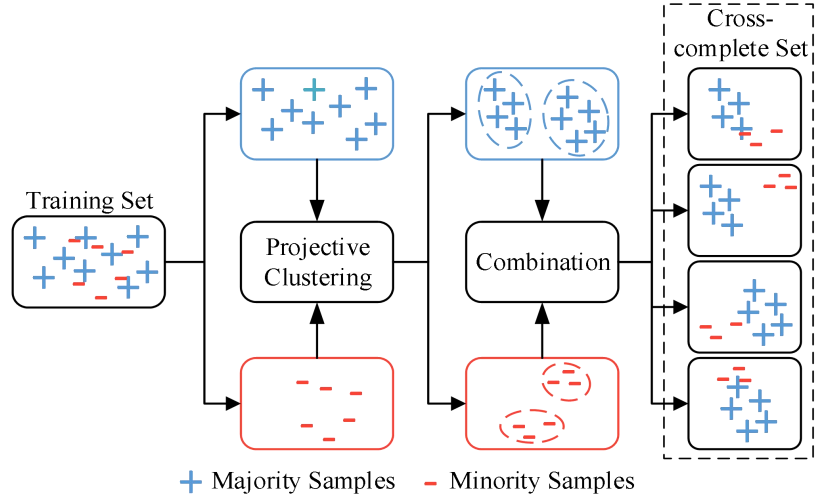

Fig. 1 The schematic diagram of PCC strategy ( $NC_{maj}$=2, $NC_{min}$=2 )

The class distribution is simple because each subset constructed by projective clustering (PC) is composed of a majority class cluster and a minority class cluster, and the samples within a cluster have similar feature distributions, so the class distribution of the subsets is relatively concentrated and simple compared with the training set. The advantage is that we only need to deal with each subset of the simple class distribution separately instead of the entire training set, thus avoiding the common problem of loosely mixed distribution that often exists in imbalanced datasets.

Intersectionality refers to the repetition of elements among subsets. Unlike the subsets constructed by the existing clustering strategies, each subset in PCC contains the same majority cluster samples or minority cluster samples as some other subsets because a majority cluster or minority cluster is assigned to different heterogeneous clusters to construct subsets multiple times in PCC. This is the core of our algorithm that can learn as much information about its samples as possible while avoiding the negative impact of overlapping majority samples in the training set. The relevant mechanism is that there is a phenomenon in which "samples in the overlapping regions of some subsets are located in the non-overlapping regions of other subsets" in the cross-complete set constructed by PCC. For the former, it is better to eliminate the overlapped samples; for the latter, it is helpful for classification to retain the samples of non-overlapping regions.

Completeness means that the CCS contains all sample information of the complete training set. At the same time, since each subset has a different class distribution, each weak classifier trained by it has a different prediction ability for different test samples. Completeness allows these weak classifiers to make comprehensive predictions for test samples from multiple perspectives, thus ensuring high prediction accuracy of the final fusion.

It is important to note that, unlike some existing clustering strategies in constructing balanced subsets, the subsets constructed by PCC are imbalanced. The PCC aims to avoid the negative impact of the imbalanced datasets on the classification caused by the loosely mixed distribution problem and to make better use of the overlapping majority sample information in the training set with the subsequent stage-wise hybrid sampling algorithm.

## 3.2 Stagewise hybrid sampling (SHS)

By constructing CCS through PCC strategy proposed in the previous section, the problem of loose mixing of distributions in imbalanced datasets can be solved effectively, making the class distribution within the subsets more uniform, but the problems of class imbalance and class overlapping in each subset have not been solved. Therefore, this paper proposes the SHS method to further solve the two problems above. First, a local overlapping region dynamic search method based on neighbors is proposed to search all local overlapping regions in each subset and undersample the majority samples in the region. Then, an iterative filtering oversampling algorithm (IFO) is designed. After reidentifying the minority class of the undersampled subset, interpolation is performed in its internal security region until the subset becomes balanced, thus further solving the problem of class imbalance and class overlapping of the subset.

The main reason for reidentifying the majority and minority classes in the undersampled subsets is that for subsets with severe class overlapping, the number of majority samples in the undersampling stage will significantly be reduced or even less than the minority samples, in which case the subsequent oversampling is performed for the original majority class. Meanwhile, LORDS sets the condition to prevent overidentification of overlapping regions when identifying it so that the number of undersampled majority samples is as few as possible and the redundancy of corresponding sample information overlap is high. Therefore, it is ensured that the original sample information of the dataset is retained to the maximum extent while eliminating the overlapping regions.

### 3.2.1 Under-sampling based on LORDS

To identify the overlapping regions more accurately, this paper proposes a local overlapping region dynamic search method (LORDS) based on neighbors. Based on the assumption that noisy samples and local boundary samples are part of the overlapping samples, this method first finds the noisy samples and local boundary samples in the dataset and uses them as the starting point of the search for overlapping regions. Under the condition of not overidentifying the overlapping regions, it iteratively searches for the remaining non-overlapping samples that satisfy the overlapping conditions to join the overlapping regions and dynamically updates the overlapping regions and non-overlapping regions. The algorithm is not limited by the distribution of the dataset and can search for local overlapping regions of arbitrary shape and distribution.

For a certain CCS subset $\mathbf{S} = \{(s_i, y_i, \hat{y}_i)\}_{i=1}^n$, where $s_i$ is an arbitrary sample in the class label $y_i \in \{g_1, g_2\}$, the overlap label $\hat{y}_i \in \{l_1, l_2\}$, $l_1$ and $l_2$ denote the corresponding samples as non-overlapping and overlapping, respectively.

Noisy samples refer to samples that most of its $R_1$ nearest neighbor samples are different from their own class[44], i. e., samples with high overlap surrounded by heterogeneous samples. Given the number $R_1$ of nearest neighbors, the noise samples $\mathbf{S}_{noise}$ can be expressed as:

$$\mathbf{S}_{noise} = \{s \mid y = g_1 \wedge \xi(y_{n_1}, y_{n_2}, ..., y_{R_1}) = g_2\} \tag{8}$$

where $\xi(\cdot)$ denotes the plurality operation and $y_{nei_n}, n = 1, 2, ..., R_1$ denotes the class label of the $n$-th nearest neighbor sample of $s_i$.

Based on the assumption that all samples whose distributions are close to heterogeneous samples are overlapping samples, the concept of local boundary samples is proposed in this paper. Suppose $s_j$ and $s_k$ are an arbitrary sample whose categories belong to $G_1$ and $G_2$, respectively. The local boundary samples $\mathbf{S}_{lb}$ is defined as:

$$\mathbf{S}_{lb} = \left\{ s_j, s_k \mid j = \arg\min_{i, i \neq k} \|s_k - s_i\|^2 \vee k = \arg\min_{i, i \neq j} \|s_j - s_i\|^2 \right\} \tag{9}$$

As Eq. (12) shows, as long as one sample is the nearest neighbor of another sample of different classes in $\mathbf{S}$, these two samples are local boundary samples.

$\mathbf{S}_{noise}$ and $\mathbf{S}_{lb}$ usually contain duplicate samples, so the initial overlap sample $\mathbf{S}_{ol}$ is defined by:

$$\mathbf{S}_{ol} = \partial(\mathbf{S}_{noise} \cup \mathbf{S}_{lb}) \tag{10}$$

where $\partial(\cdot)$ denotes the sample deduplication operation.

From the description of the above two samples, it is easy to see that noisy samples and local boundary samples are both samples whose distribution is closer to that of the heterogeneous samples in the dataset, i. e., samples with higher class overlapping, so the assumption that noisy samples and local boundary samples are part of the overlapping samples is reasonable.

After determining the initial overlapping samples $\mathbf{S}_{ol}$, LORDS then finding the $R_2$ nearest neighbors of all the remaining non-overlapping samples in $\mathbf{S}$ and identifies the potential overlapping samples $\mathbf{P}$ according to Eq. (11).

$$\mathbf{P} = \{s_i \in \mathbf{S} \setminus \mathbf{S}_{ol} \mid \hat{y}_i = l_1 \vee \xi(\hat{y}_{nei_1}, \hat{y}_{nei_2}, ..., \hat{y}_{nei_{R_2}}) = l_2\} \tag{11}$$

where $\hat{y}_{nei_n}, n = 1, 2, ..., R_2$ denotes the overlapping label of the $n$-th nearest neighbor sample

$s_{nei_n}, n=1,2,...,R_2$ of $s_t$. In other words, if most of the $R_2$-nearest neighbors of a non-overlapping sample are overlapping samples, it will be judged as a potential overlapping sample $s_p$. For $s_p$, LORDS will find all overlapping samples $\mathbf{O}$ in its $R_2$-nearest neighbors as follows:

$$\mathbf{O} = \{s_{nei_n} | \hat{y}_{nei_n} = l_2, s_p \in \mathbf{P}\} \tag{12}$$

Then, the distance $d_1$ from $s_p$ to the center $\mu_o$ of $\mathbf{O}$ and the average distance $d_2$ from all samples in $\mathbf{O}$ to $\mu_o$ are calculated as follows:

$$d_1 = \|s_p - \mu_o\|^2 \tag{13}$$

$$d_2 = \frac{1}{N_O} \sum_{s_i \in \mathbf{O}} \|s_i - \mu_o\|^2 \tag{14}$$

where $N_O$ is the total number of samples in $\mathbf{O}$. If the following condition (18) is met, the potential overlapping samples $s_p$ will be judged as an overlapping sample; otherwise, it will still be judged as non-overlapping samples and continue to participate in the next round of searching.

$$d_1 \leq d_2 \vee N_O = R_2 \tag{15}$$

That is, if the distance between a potential overlapping sample $s_p$ and the center of overlapping samples in its $R_2$-nearest neighbors is not greater than the average distance from these overlapping samples to their center, or all of its $R_2$-nearest neighbor samples are overlapping, $s_p$ will be determined to be overlapping sample. This condition is set to prevent overidentifying the overlapping region and thus avoid removing useful non-overlapping region majority samples in SHS.

For the overlapping regions of each subset identified by the LORDS, the overlapping majority samples will be removed and sent to IFO for balancing by synthesizing minority samples. It is important to note that for datasets with particularly mixed class distributions, there may be cases where all the majority class samples in the data overlap, i.e., flag equals to 1. In this case, in order not to lose excessive sample information, all overlapping majority class samples in the data will be retained and the subsequent oversampling step will be performed directly.

**3.2.2 Oversampling based on IFO**

The under-sampling operation of overlapping majority samples in each subset not only eliminates the class overlapping problem but also alleviates the class imbalance problem to a certain extent. To further solve the class imbalance problem that still exists after the subsets are deoverlapped, a simple iterative filtering oversampling algorithm is designed in this paper for balancing the under-sampled subsets.

For the subset with a high degree of class overlapping, the number of majority sample in the under-sampling stage will be significantly reduced or even less than the minority. Therefore, IFO first reidentifies the majority and minority classes and then iteratively generates new minority samples using the SMOTE [10] algorithm and uses ENN [44] screening to ensure that the generated samples are located in a safe region within the minority class.

Assuming that $\tilde{\mathbf{S}}$ is the dataset after subset $\mathbf{S}$ undersamples the overlapping majority samples, $\tilde{\mathbf{S}}_{min}$ is its subset of minority samples and $s_i^{min}$ is an arbitrary sample in $\tilde{\mathbf{S}}_{min}$. IFO first calculates the Euclidean distance from $s_i^{min}$ to all other samples in $\tilde{\mathbf{S}}_{min}$ to obtain its $R_3$ nearest neighbor samples $\mathbf{S}_{nei} = \{s_{nei_n}\}, n = 1, 2, ..., R_3$. Then, a sample $s_{nei_j}, j \in [1, R_3]$ is randomly selected from its $R_3$-nearest neighbors and used to construct a new minority sample $s_{new}$ according to Eq. (16) with $x_{mi}$.

$$s_{new} = s_i^{min} + \alpha \cdot (s_{nei_j} - s_i^{min}) \tag{16}$$

where $\alpha$ is a random number in $(0,1)$. Then, we calculate the Euclidean distances from $s_{new}$ to all samples in $\tilde{\mathbf{S}}$ to obtain its $R_3$-nearest neighbors. If most of $R_3$-nearest neighbor samples of $s_{new}$ belong to the minority class, it will be retained, and then the minority sample subset $\tilde{\mathbf{S}}_{min}$ and the total dataset $\tilde{\mathbf{S}}$ needs to be updated as follows:

$$\tilde{\mathbf{S}}_{min} = \tilde{\mathbf{S}}_{min} \cup s_{new} \tag{17}$$

$$\tilde{\mathbf{S}} = \tilde{\mathbf{S}} \cup s_{new} \tag{18}$$

Subsequent oversampling will be performed based on the updated dataset until $\tilde{\mathbf{S}}$ is fully balanced, i. e., the number of the minority equals that of the majority.

### 3.3 Projective clustering transfer mapping mechanism (CTM)

Since the overlap identification process in this paper is based on Euclidean space, and the Euclidean distance is not an accurate distance measure in the high-dimensional case, the subset may still have a certain class overlapping problem after SHS. In order to further reduce the class overlapping of subsets after SHS and improve the richness of single-sample information, we introduce the transfer learning approach to design an envelope clustering transfer mapping mechanism for assisting the training and prediction of the corresponding weak classifier.

Specifically for each SHS subset, the process of envelope clustering transfer mapping mechanism in this paper is set as follows: first, according to the given clustering ratio, the two classes of samples in the subset (original subset) are clustered separately using the K-means method, and all the cluster centers form the new samples with low overlap (updated subset); second, the domain transfer process is conducted with the updated subset as the source domain and the SHS subset as the target domain, and the updated subset is updated again to be the common domain (final CTM layer subset). The transfer process comprehensively considers the locality and globality of the sample space. The locality is reflected in the local preservation projection algorithm used for dimensionality reduction, and the globality is reflected in that the dimensionality reduction process is constrained by the global maximum mean discrepancy, so that the samples of the CTM subset have the advantages of rich sample information and low class overlapping in the clustering layer, while the distribution is close to that of the SHS subset. The schematic diagram of the CTM is shown in Fig. 2.

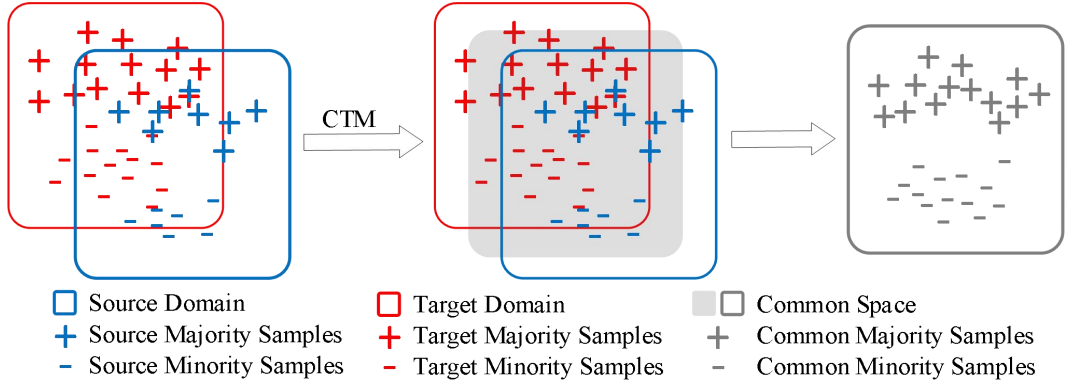

Fig. 2 The schematic diagram of CTM mechanism

The mechanism works as follows: First, consider an extreme case where the majority and minority classes of the data are clustered into one cluster, respectively, and the two cluster centers form a post-clustering subset. At this point, the sample has the best separability and the lowest class overlapping. Therefore, it is easy to understand that as the ratio of the number of clusters to the original number of samples gradually decreases, the class overlapping of the clustered SHS subsets will gradually decrease and the separability will gradually increase. But at the same time, more sample information will be lost, so the clustering ratio should not be set too small.

In this section, the subscripts $S$ and $T$ represent the relevant parameters of the source domain and target domain, respectively. $\mathbf{D}$ represents a sample space with a marginal sample distribution $P(\mathbf{d})$, $\mathbf{d}$ represents a sample, the source data (subset after clustering) is expressed as $\mathbf{D}_S = [\mathbf{d}_{S_1}, \mathbf{d}_{S_2}, ..., \mathbf{d}_{S_{N_S}}]^T$, the target data (subset before clustering) is expressed as $\mathbf{D}_T = [\mathbf{d}_{T_1}, \mathbf{d}_{T_2}, ..., \mathbf{d}_{T_{N_T}}]^T$, $N_S$ and $N_T$ are the number of samples in the source and target domains, respectively. The label vector of the sample subset is expressed as $\mathbf{Y}_S = [y_{S_1}, y_{S_2}, ..., y_{S_{N_S}}]^T$. C is the number of classes, the symbol $\|\cdot\|_H$ is the reproducing kernel Hilbert space norm, $\tau(\cdot)$ denotes the trace operator and $N_{K(\cdot)}$ denotes the $k$ nearest neighbors operator.

### 3.3.1 Marginal distribution adaptation

Suppose there is a transformation $\varphi(\cdot)$ such that $P(\varphi(\mathbf{D}_S)) \approx P(\varphi(\mathbf{D}_T))$. As a commonly used criterion for measuring differences between distributions, MMD is widely used in transfer learning, domain adaptation and other fields. The distance between the source data (subset after clustering) and the target data (subset before clustering) can be written empirically through MMD as follows[45]:

$$dist(\mathbf{D}_S, \mathbf{D}_T) = \left\| \frac{1}{n_1} \sum_{i=1}^{n_1} \varphi(\mathbf{d}_{S_i}) - \frac{1}{n_2} \sum_{j=1}^{n_2} \varphi(\mathbf{d}_{T_j}) \right\|_H^2 \quad (19)$$

The nonlinear mapping $\varphi(\cdot)$ can be found by minimizing the quantity. However, directly optimizing the mapping and the number may fall into the local minimum, which is very difficult to solve. According to the unsupervised dimensionality reduction method MMDE [46], we can embed both the source domain and the target domain into a common low-dimensional space by learning the kernel matrix $\mathbf{K}$. The kernel mapping can be considered as: $\mathbf{D} \to \varphi(\mathbf{D}) = [\varphi(\mathbf{d}_1), \varphi(\mathbf{d}_2), ..., \varphi(\mathbf{d}_n)]$ and $\mathbf{K} = \varphi(\mathbf{D})^T \varphi(\mathbf{D})$. Specifically, after mapping, $\mathbf{D}_S$ and $\mathbf{D}_T$ can be written as: $\begin{bmatrix} \varphi(\mathbf{D}_S)\varphi(\mathbf{D}_S) & \varphi(\mathbf{D}_S)\varphi(\mathbf{D}_T) \\ \varphi(\mathbf{D}_T)\varphi(\mathbf{D}_S) & \varphi(\mathbf{D}_T)\varphi(\mathbf{D}_T) \end{bmatrix}$, thereby, $\mathbf{K} = \begin{bmatrix} \mathbf{K}_{S,S} & \mathbf{K}_{S,T} \\ \mathbf{K}_{T,S} & \mathbf{K}_{T,T} \end{bmatrix}$.

In terms of trace operation trick, the distance between samples in source domain and target domain is equivalent to $\tau(\mathbf{KM})$, and subject to constraints on $\mathbf{K}$. $\mathbf{M}$ is MMD matrices.

### 3.3.2 Local structure preservation

However, the reduction of difference in the marginal distributions may lead to the destruction of the relationship structure between samples and the loss of useful information. Therefore, the use of affinity matrix to preserve neighborhood structure can be considered. Under the manifold assumption, a dimensionality reduction algorithm called LPP [47] aims to optimally preserve the neighborhood structure of data. Its objective function can be expressed as:

$$\sum_{i,j} (y_i - y_j)^2 A_{ij} \tag{20}$$

where $\mathbf{Y} = [y_1, y_2, ..., y_n]^T$ is the map of $\mathbf{D} = [d_1, d_2, ..., d_n]^T$, Affinity matrix $\mathbf{A}$ can be calculated in the following two ways.

Simple-minded:

$$A_{ij} = \begin{cases} 1, & \text{if } d_i \in \mathbf{N}_{K(d_j)} \| d_j \in \mathbf{N}_{K(d_i)} \\ 0, & \text{others} \end{cases} \tag{21}$$

Heat-kernel:

$$A_{ij} = \begin{cases} e^{-\frac{\|d_i - d_j\|^2}{f}}, & \text{if } d_i \in \mathbf{N}_{K(d_j)} \| d_j \in \mathbf{N}_{K(d_i)} \\ 0, & \text{others} \end{cases} \tag{22}$$

where $f$ is the kernel parameter. $A_{ij}$ will be assigned a large value if $d_i$ is the neighborhood of $d_j$. Based on this idea, the neighborhood structure preservation in this paper can be defined as the following formula, so as to better maintain the neighborhood relationship between source domain and target domain samples.

$$\frac{1}{2} \sum_{m,n} \|\varphi(d_{S_m}) - \varphi(d_{S_n})\|_H^2 A_{mn} + \frac{1}{2} \sum_{p,q} \|\varphi(d_{T_p}) - \varphi(d_{T_q})\|_H^2 A_{pq} \tag{23}$$

where $\varphi(\mathbf{D}_S) = [\varphi(d_{S_1}), \varphi(d_{S_2}), ..., \varphi(d_{S_m})]^T$ is the map of samples in source domain, and $\varphi(\mathbf{D}_T) = [\varphi(d_{T_1}), \varphi(d_{T_2}), ..., \varphi(d_{T_p})]^T$ is the map of samples in target domain. $A_{mn}$ and $A_{pq}$ are the elements of affinity matrix for source and target domains, respectively.

### 3.3.3 Joint optimization

This method aims to align the distribution of source domain (post cluster subset) and target domain (pre cluster subset) under the condition of preserving domain structure. The joint local structure distribution alignment term can be shown as follows.

$$\begin{aligned}
\min\ & \tau\left(\frac{1}{n_S^2}\varphi(\mathbf{D}_S)II^{\mathrm{T}}\varphi^{\mathrm{T}}(\mathbf{D}_S) + \frac{1}{n_T^2}\varphi(\mathbf{D}_T)II^{\mathrm{T}}\varphi^{\mathrm{T}}(\mathbf{D}_T)\right.\\
&\left. - \frac{1}{n_S n_T}\varphi(\mathbf{D}_S)II^{\mathrm{T}}\varphi^{\mathrm{T}}(\mathbf{D}_T) - \frac{1}{n_S n_T}\varphi(\mathbf{D}_S)I^{\mathrm{T}}I\varphi(\mathbf{D}_T)\right)\\
& + \frac{1}{2}\sum_{m,n}\tau\left(\varphi(d_{S_m})\varphi^{\mathrm{T}}(d_{S_m}) + \varphi(d_{S_n})\varphi^{\mathrm{T}}(d_{S_n})\right.\\
&\left. - \varphi(d_{S_m})\varphi^{\mathrm{T}}(d_{S_n}) - \varphi(d_{S_n})\varphi^{\mathrm{T}}(d_{S_m})\right)A_{mn}\\
& + \frac{1}{2}\sum_{p,q}\tau\left(\varphi(d_{T_p})\varphi^{\mathrm{T}}(d_{T_p}) + \varphi(d_{T_q})\varphi^{\mathrm{T}}(d_{T_q})\right.\\
&\left. - \varphi(d_{T_p})\varphi^{\mathrm{T}}(d_{T_q}) - \varphi(d_{T_q})\varphi^{\mathrm{T}}(d_{T_p})\right)A_{pq}
\end{aligned} \quad (24)$$

Combining the properties of matrix trace and our previous definition, Eq. (24) can be simplified into Eq. (25).

$$\min\ \tau(\hat{\mathbf{K}}\hat{\mathbf{M}}) + \tau\left(\varphi(\mathbf{D}_S)\varphi^{\mathrm{T}}(\mathbf{D}_S)\hat{\mathbf{L}}_S\right) + \tau\left(\varphi(\mathbf{D}_T)\varphi^{\mathrm{T}}(\mathbf{D}_T)\hat{\mathbf{L}}_T\right) \quad (25)$$

where $\hat{\mathbf{K}}$ is the kernel matrix and $\hat{\mathbf{M}}$ is the MMD matrices, both are obtained from samples after first transfer step. $\varphi(\mathbf{D}_S)\varphi^{\mathrm{T}}(\mathbf{D}_S)$ is $\hat{\mathbf{K}}_S$ and $\varphi(\mathbf{D}_T)\varphi^{\mathrm{T}}(\mathbf{D}_T)$ is $\hat{\mathbf{K}}_T$. $\hat{\mathbf{L}}_S = \hat{\mathbf{D}}_S - \hat{\mathbf{A}}_S$ and $\hat{\mathbf{L}}_T = \hat{\mathbf{D}}_T - \hat{\mathbf{A}}_T$ are the Laplacian matrixes of source domain and target domain, $\hat{\mathbf{D}}_{mm} = \sum_n \hat{A}_{mn}$ and $\hat{\mathbf{D}}_{qq} = \sum_p \hat{A}_{pq}$ are both diagonal matrixes, $\hat{\mathbf{A}}$ is affinity matrix. To simplify the calculation, we can simplify Eq. (25) to Eq. (26):

$$\min\ \tau(\hat{\mathbf{K}}\hat{\mathbf{M}}) + \tau(\hat{\mathbf{K}}\cdot\hat{\mathbf{L}}) \quad (26)$$

where $(\cdot)$ denotes dot multiplication of $\hat{\mathbf{K}}$ and $\hat{\mathbf{L}}$, $\hat{\mathbf{L}} = \begin{bmatrix}\hat{\mathbf{L}}_S & 0 \\ 0 & \hat{\mathbf{L}}_T\end{bmatrix}$.

Due to high computational cost of MMDE, a unified kernel learning method is adopted which utilizes an explicit low-rank representation. Thus, Eq. (27) can be acquired.

$$\min_{\mathbf{W}}\ \tau(\mathbf{W}^{\mathrm{T}}\hat{\mathbf{K}}\hat{\mathbf{M}}\hat{\mathbf{K}}\mathbf{W}) + \tau(\mathbf{W}^{\mathrm{T}}\hat{\mathbf{L}}^*\mathbf{W}) + \lambda\tau(\mathbf{W}^{\mathrm{T}}\mathbf{W}) \\ \text{s.t.}\ \mathbf{W}^{\mathrm{T}}\hat{\mathbf{K}}\hat{\mathbf{D}}\hat{\mathbf{K}}\mathbf{W} = \mathbf{I}_m \quad (27)$$

The pseudo code of the CTM algorithm is shown in the Algorithm 4.

### 3.4 The overall algorithm

In terms of the characteristics of class imbalance and class overlapping existing in imbalanced datasets, this paper proposes an ensemble learning method based on dual clustering and stage-wise hybrid sampling. The proposed algorithm can solve the problems of class imbalance and class overlapping at the same time, and mainly includes three parts: PCC, SHS, and CTM. The flow chart of the proposed algorithm is shown in Fig. 5. First, an imbalanced and overlapping cross-complete set is constructed by PCC on the training set. Then, SHS is used to de-overlap and balance the subsets in the dataset to obtain a set of subsets balanced and low overlapping. Thirdly, a clustering mapping layer with lower class overlapping and richer single-sample information is constructed to obtain high quality of new samples with better balancing and low overlapping. Finally, the final prediction result is obtained by decision level fusion.

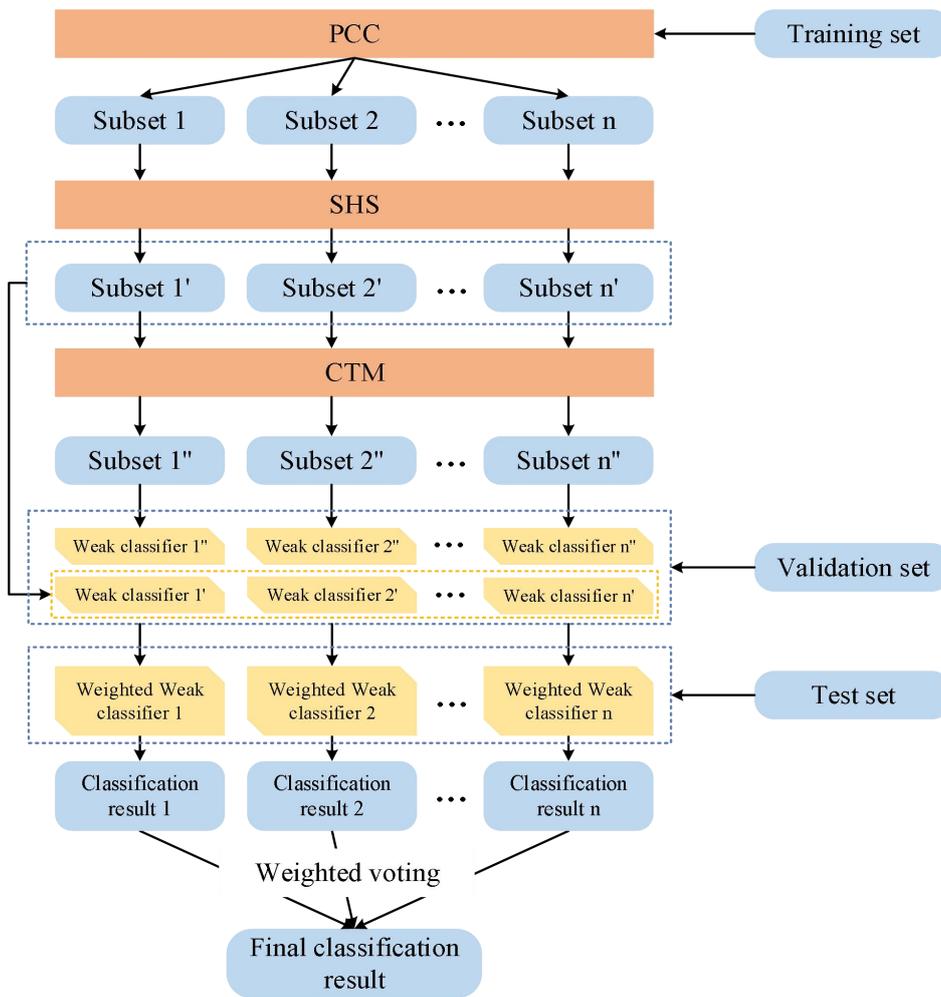

Fig. 3 The flow chart of the overall algorithm.

# 4 Experimental results and analysis

Note: the results here is part of the full results. Please contact the corresponding author for full results.

## 4.1 Experimental condition
### 4.1.1 Experimental settings

In this paper, three groups of experiments are conducted, and they are: verification of the proposed algorithm, algorithms comparison, and diversity analysis.

For fair comparison, in this paper, the widely used SVM is used as the base classifier. The results of all datasets are obtained through 10 rounds of 5-fold cross-validation experiments. The values of all classification performance metrics are taken from the arithmetic mean of these 50 experiments.

Since the DCSHS algorithm in this paper involves resampling and ensemble learning, classical method and state of the art methods on the existing oversampling methods, undersampling methods, hybrid sampling methods and ensemble learning methods in the field of imbalanced learning are selected, respectively. Among the 12 methods, SMOTE [10], KNSMOTE[14] and ADPCHFO[48] belong to oversampling methods, ENN[44], CBIS[20], and AdaOBU[5] belong to undersampling methods, SMTL[21], CUSS[49], and RFMSE[24] belong to hybrid sampling methods, and SBE[50], DTE-SBD[31], REMDD[33], and EASE[51] belong to ensemble sampling methods. The four kinds of methods are called m1, m2, m3 and m4. Some of them are published in recent three years.

**4.1.2 Data**

In this paper, 39 widely used publicly available imbalanced datasets from the UCI and KEEL public databases were selected for method validation. To objectively evaluate the classification performance of the algorithms, the datasets were selected to ensure their diversity and representativeness as much as possible. The final datasets were selected from different fields, such as medicine, biology, and physics, with the number of samples ranging from 101 to 19871, the number of features ranging from 3 to 41, and the imbalance ratio ranging from 1.79 to 763.27. According to the IR, the datasets can be divided into two parts: datasets with low and high IR, datasets with higher IR. Table 1 lists the main information of each dataset, including the name, total number of samples, and imbalance ratio (IR). IR is the ratio of the sample size of the majority and minority classes in the data.

Table 1 The properties of datasets

| Dataset | Instances | Features | IR | Dataset | Instances | Features | IR |
|---|---|---|---|---|---|---|---|
| **Datasets with low and high IR:** | | | | | | | |
| ionosphere | 351 | 32 | 1.79 | glass06vs5 | 108 | 9 | 11.00 |
| glass1 | 214 | 9 | 1.82 | ecoli01vs5 | 240 | 6 | 11.00 |
| Monk | 601 | 6 | 1.92 | glass0146vs2 | 205 | 9 | 11.06 |
| haberman | 306 | 3 | 2.78 | glass2 | 214 | 9 | 11.59 |
| Vehicle1 | 846 | 18 | 2.85 | ecoli0146vs5 | 280 | 6 | 13.00 |
| Vehicle2 | 846 | 18 | 2.88 | yeast1vs7 | 459 | 8 | 14.30 |
| Vehicle3 | 846 | 18 | 2.99 | glass4 | 214 | 9 | 15.46 |
| ecoli1 | 336 | 7 | 3.36 | ecoli4 | 336 | 7 | 15.80 |
| yeast2 | 1484 | 8 | 5.08 | zoo3 | 101 | 16 | 19.20 |
| ecoli2 | 336 | 7 | 5.46 | Lymphographynormalfibrosis | 148 | 18 | 23.67 |
| yeast3 | 1484 | 8 | 8.10 | Winequalityred4 | 1599 | 11 | 29.17 |
| ecoli3 | 336 | 7 | 8.60 | poker9vs7 | 244 | 10 | 29.50 |
| yeast2vs4 | 514 | 8 | 9.08 | abalone3vs11 | 502 | 8 | 32.47 |
| ecoli067vs35 | 222 | 7 | 9.09 | winequalitywhite9vs4 | 168 | 11 | 32.60 |
| ecoli0234vs5 | 202 | 7 | 9.10 | yeast6 | 1484 | 8 | 41.40 |
| yeast0359vs78 | 506 | 8 | 9.12 | winequalitywhite39vs5 | 1482 | 11 | 58.28 |
| yeast02579vs368 | 1004 | 8 | 9.14 | shuttle2vs5 | 3316 | 9 | 66.67 |
| ecoli01vs235 | 244 | 7 | 9.17 | poker8vs6 | 1477 | 10 | 85.88 |
| **Datasets with higher IR:** | | | | Abalone19 | 4174 | 8 | 129.4 |
| kddrootkitback | 2225 | 41 | 100.14 | cod | 19871 | 8 | 763.27 |

**4.1.3 Experimental configuration**

The experiments use a 64-bit Windows 10 computer and the hardware parameters of the experiment platform are CPU (Intel i5-8400), 8 GB memory. The experiments run on Matlab R2018b. The set of parameters in this paper is as follows. In the PCC step, the number of clusters for both majority and minority classes is 1 to 3. In the SHS step, the number of nearest neighbors $R_1$, $R_2$ and $R_3$ are all 5. In the CTM step, the number of clusters is 0.5 times the number of samples in the subset, the regularization parameter lambda is 0.01, kernel type is 'rbf', the bandwidth for rbf kernel' gamma is 100, and the affinity matrix mode is 'simple' mode.

**4.2 Algorithms comparison**

In this section, the performance of the proposed method is investigated on datasets with low IR, high IR and higher IR imbalanced datasets, respectively.

**4.2.1 Comparison of datasets with low and high IR**

In this section, the experimental results are presented and analyzed in detail. Table 3 gives the detailed classification results for all compared algorithms on the datasets with low and high IR. Due to the size limitation of the table, the recall metric is abbreviated as Rec, the F1-measure metric is abbreviated as F1-M, and the G-mean metric is abbreviated as G-M. The best performance on evaluation metric value is identified in bold black. Assuming the first rank for the method with the best performance and the fourteenth rank for the method with the worst performance, Table3 also gives the rank of the algorithms. Every value is shown as (value rank).

Table 2 Classification performance of different algorithms using SVM on the datasets with low and high IR. The superscript m1-m4 represents the methods of oversampling, undersampling, hybrid sampling and ensemble learning (part of results)

| Dataset | Ionosphere | | | | glass1 | | | | Monk | | | |
|---|---|---|---|---|---|---|---|---|---|---|---|---|
| Measure | Rec | F1-M | G-M | AUC | Rec | F1-M | G-M | AUC | Rec | F1-M | G-M | AUC |
| SMOTE[m1] | 0.700 6 | 0.747 10 | 0.793 9 | 0.802 9 | 0.909 3 | 0.560 3 | 0.486 9 | 0.587 5 | **0.847** **1** | 0.577 2 | 0.606 3 | 0.642 2 |
| KNSMOTE[m1] | 0.664 12 | 0.741 12 | 0.784 12 | 0.797 12 | 0.235 11 | 0.281 12 | 0.405 12 | 0.541 12 | 0.528 11 | 0.424 10 | 0.511 10 | 0.514 11 |
| ADPCHFO[m1] | 0.657 13 | 0.747 9 | 0.786 11 | 0.801 10 | 0.130 14 | 0.190 14 | 0.323 14 | 0.528 13 | 0.644 9 | 0.486 9 | 0.555 9 | 0.565 9 |
| ENN[m2] | 0.642 14 | 0.735 13 | 0.777 13 | 0.793 13 | 0.225 13 | 0.298 11 | 0.426 11 | 0.548 11 | 0.024 14 | 0.046 14 | 0.154 14 | 0.512 12 |
| CBIS[m2] | 0.762 2 | 0.754 5 | 0.806 3 | 0.809 4 | 0.950 2 | 0.560 4 | 0.433 10 | 0.578 8 | 0.669 7 | 0.524 7 | 0.600 6 | 0.607 6 |
| AdaOBU[m2] | 0.740 4 | 0.637 14 | 0.704 14 | 0.707 14 | 0.232 12 | 0.234 13 | 0.363 13 | 0.425 14 | 0.540 10 | 0.415 11 | 0.492 11 | 0.501 14 |
| SMTL[m3] | 0.717 5 | 0.753 7 | 0.800 4 | 0.807 5 | 0.905 4 | 0.560 5 | 0.487 8 | 0.587 6 | 0.833 3 | 0.566 3 | 0.595 7 | 0.632 3 |
| CUSS[m3] | 0.695 7 | 0.754 6 | 0.796 7 | 0.807 7 | 0.607 10 | 0.481 10 | 0.503 4 | 0.592 3 | 0.404 12 | 0.260 12 | 0.241 13 | 0.537 10 |
| RFMSE[m3] | 0.688 10 | 0.759 3 | 0.799 5 | 0.810 3 | 0.832 7 | 0.540 8 | 0.503 5 | 0.572 10 | 0.656 8 | 0.491 8 | 0.558 8 | 0.571 8 |
| SBE[m4] | 0.676 11 | 0.742 10 | 0.786 10 | 0.798 11 | 0.610 9 | 0.483 9 | 0.535 3 | 0.577 9 | 0.201 13 | 0.196 13 | 0.308 12 | 0.506 13 |
| DTE-SBD[m4] | 0.690 9 | 0.753 8 | 0.796 8 | 0.806 8 | 0.892 5 | 0.557 6 | 0.491 7 | 0.585 7 | 0.780 4 | 0.556 5 | 0.603 5 | 0.625 5 |
| REMDD[m4] | 0.691 8 | 0.755 4 | 0.797 6 | 0.807 6 | 0.878 6 | 0.555 4 | 0.502 6 | 0.588 4 | 0.769 5 | 0.555 6 | 0.606 2 | 0.626 4 |
| EASE[m4] | **0.832** **1** | 0.761 2 | 0.813 2 | 0.815 2 | 0.749 8 | 0.654 2 | 0.627 2 | 0.637 2 | 0.674 6 | 0.562 4 | 0.604 4 | 0.605 7 |
| **DCSHS** | 0.742 3 | **0.810** **1** | **0.834** **1** | **0.840** **1** | **0.957** **1** | **0.611** **1** | **0.572** **1** | **0.649** **1** | 0.839 2 | **0.626** **1** | **0.684** **1** | **0.701** **1** |

In addition, to test the significance level of the differences in rankings between the different methods, we performed Holm's p-value test on the performance of the rankings of DCSHS and other algorithms on the four categorical evaluation criteria. The results are presented in Table 3. For the results of Holm test, it is generally considered that a p-value less than the threshold value of 0.05 indicates that the two subjects in the test differ significantly, i. e., not by chance.

Table 3 The *p*-value of Holm test with DCSHS as control method.

| Algorithms | Rec | F1-M | G-M | AUC |
|---|---|---|---|---|
| SMOTE | 1.42E-03 | 3.97E-06 | 1.30E-03 | 2.21E-03 |
| KNSMOTE | 9.78E-20 | 1.16E-12 | 2.53E-20 | 1.85E-20 |
| ADPCHFO | 2.49E-21 | 6.32E-02 | 1.47E-15 | 3.11E-10 |
| ENN | 1.91E-45 | 7.89E-16 | 6.35E-50 | 1.04E-40 |
| CBIS | 4.83E-29 | 3.13E-05 | 2.92E-32 | 8.69E-25 |
| AdaOBU | 9.47E-10 | 3.59E-20 | 1.96E-32 | 2.65E-28 |
| SMTL | 9.98E-02 | 4.53E-08 | 1.50E-03 | 4.12E-03 |
| CUSS | 2.59E-30 | 3.97E-06 | 6.48E-32 | 1.86E-23 |
| RFMSE | 2.13E-01 | 7.42E-27 | 1.98E-14 | 4.22E-17 |
| SBE | 5.42E-01 | 2.54E-21 | 1.13E-10 | 9.41E-11 |
| DTE-SBD | 1.22E-03 | 2.36E-12 | 3.59E-09 | 4.86E-09 |
| REMDD | 1.29E-02 | 1.03E-15 | 6.21E-12 | 2.48E-09 |
| EASE | 1.97E-01 | 1.59E-16 | 4.83E-17 | 3.78E-18 |

In general, from Tables 1, 3, we can know that for different IR datasets, DCSHS perform overall better than other comparison algorithms. Thanks to the intersectionality of the subsets constructed by PCC, DCSHS avoids the trade-off between more emphasis on boundary samples or more emphasis on security region samples for balance. The former will lead to a low G-mean, while the latter will lead to a low F1 score. Our algorithm achieves a good compromise between the two cases. Among all methods except DCSHS, SMOTE and SMTL perform well for all four metrics, which is probably due to the strong generalization performance of the classical oversampling algorithm SMOTE.

### 4.2.2 Comparison of datasets with higher IR

The proposed method DCSHS performs better on datasets with low and high IR. To further validate the effectiveness of the proposed method, this section compares the proposed method as well as the four imbalanced ensemble methods SBE, DTE-SBD, REMDD, and EASE, with the highly imbalanced dataset. Table 5 gives the detailed classification results for all compared algorithms. The best performance on evaluation metric value is identified in bold black. From Table 4, it can be seen the proposed method DCSHS is superior to other methods.

Table 4 Classification performance of different algorithms using SVM on the datasets with higher IR

| Dataset | Measure | SBE | DTE-SBD | REMDD | EASE | DCSHS |
|---|---|---|---|---|---|---|
| Abalone19 | Rec | 0.615 | 0.857 | 0.875 | 0.690 | **0.909** |
| | F1-M | 0.026 | 0.037 | 0.020 | 0.038 | **0.058** |
| | G-M | 0.565 | 0.802 | 0.634 | 0.657 | **0.819** |
| | AUC | 0.567 | 0.804 | 0.667 | 0.690 | **0.824** |

## 5  Conclusion

The study of imbalance learning problems has been a hotspot and a challenging research area in machine learning. Most of the existing algorithms focus on the class imbalance problem but seldom consider the class overlapping problem. Due to the limitation, the overlap problem is not solved even becomes worse during imbalanced learning. Besides, the core problem of how to better learn overlapping sample information has not been well solved in the existing methods.

To address the above problems, the DCSHS method is proposed in this paper to better learn class overlapping sample information and deal with class imbalance and class overlapping problems simultaneously. The process of solving the class overlapping and class imbalance problem by DCSHS is as follows: firstly, the training set with scattered and mixed distribution is converted into a set of subsets with simple CCS distribution by PCC, which can perform clustering and feature reduction simultaneously, to facilitate the subsequent processing. Then, the de-overlap and balance operation of each subset with simple class distribution is completed by SHS. Thanks to the precise overlap identification effect of LORDS, SHS can eliminate class overlapping regions by eliminating as few overlapping majority class samples as possible. Finally, a subset of cluster transfer mapping with further reduction of class overlapping and richer single sample information is constructed by CTM to assist the SHS subset. Compared with the existing related algorithms, DCSHS can achieve better learning of overlapping samples.

Although the method in this paper has achieved good results, there are still some work to do. Future work can consider the following direction. The algorithm in this paper currently searches for class overlapping regions based on Euclidean space, so other distance measures and judgment criteria can be considered for the identification of overlapping regions in the future.

**Data availability**
The data and codes can be found at: https://pan.baidu.com/s/1yT5H8rlj6JCl6323hVacIA, extraction code: 1111).

# Rreferences